\documentclass{article}

\PassOptionsToPackage{numbers, compress}{natbib}

\usepackage[preprint]{neurips_2023}
\usepackage{graphicx}
\usepackage{amsmath}
\usepackage{amssymb}
\usepackage{colortbl}
\usepackage[dvipsnames, svgnames, x11names]{xcolor}

\usepackage[utf8]{inputenc} 
\usepackage[T1]{fontenc}    
\usepackage{hyperref}       
\usepackage{url}            
\usepackage{booktabs}       
\usepackage{amsfonts}       
\usepackage{nicefrac}       
\usepackage{microtype}      
\usepackage{xcolor}         

\title{Shape Guided Gradient Voting for Domain Generalization}

\author{Jiaqi Xu$^1$\thanks{Work done during internships at Microsoft Research Asia.}
, Yuwang Wang$^2$  
, Xuejin Chen$^1$ \\
\texttt{xujiaqi@mail.ustc.edu.cn},\\ \texttt{wang-yuwang@mail.tsinghua.edu.cn},\\ \texttt{xjchen99@ustc.edu.cn}\\
$^1$University of Science and Technology of China, $^2$Tsinghua University
}

\begin{document}

\maketitle

\begin{abstract}
Domain generalization aims to address the domain shift between training and testing data. To learn the domain invariant representations, the model is usually trained on multiple domains. It has been found that the gradients of network weight relative to a specific task loss can characterize the task itself~\cite{achille2019task2vec}. In this work, with the assumption that the gradients of a specific domain samples under the classification task could also reflect the property of the domain, we propose a Shape Guided Gradient Voting (SGGV) method for domain generalization. Firstly, we introduce shape prior via extra inputs of the network to guide gradient descending towards a shape-biased direction for better generalization. Secondly, we propose a new gradient voting strategy to remove the outliers for robust optimization in the presence of shape guidance. To provide shape guidance, we add edge/sketch extracted from the training data as an explicit way, and also use texture augmented images as an implicit way. We conduct experiments on several popular domain generalization datasets in image classification task, and show that our shape guided gradient updating strategy brings significant improvement of the generalization.
\end{abstract}

\section{Introduction}
\label{sec:intro}

In recent years, deep learning has made great progress in many research fields, such as computer vision, natural language processing, speech, robotics and so on. 
Usually, the training and testing data are assumed to be sampled from the same data distribution. In practical application, however, this assumption is often hard to meet in practice, which results in the domain discrepancy between training and testing data. To address this issue, domain generalization task aims to improve the performance of the model trained on source domains when tested on unseen target domain directly without any adaptation. 

The previous domain generalization methods have mainly focused on $(i)$ learning domain invariant features~\cite{muandet2013domain,ghifary2015domain,li2018domain,wang2018learning,kim2021selfreg,Chen_2021_ICCV},  $(ii)$ adding or learning domain-invariant knowledge, including meta-learning for regularization~\cite{balaji2018metareg}, shape prior~\cite{carlucci2019domain,shi2020informative,Narayanan_2021_ICCV}, self-challenging~\cite{huangRSC2020}, Fourier space~\cite{Huang_2021_CVPR,Xu_2021_CVPR}, and $(iii)$ augmentation~\cite{zhou2021domain,Li_2021_cvpr,Li_2021_CVPR_progressive, volpi2018generalizing} to increase the diversity of training data and expand the supporting space. Despite great efforts for various solutions, some of the essential settings for domain generalization are still less explored. For example, recent work~\cite{gulrajani2020search,cha2021swad} has made great progress in model selection for domain generalization task, which is an essential setting for machine learning. In this paper, we focus on another essential part, gradient. Gradient has been found to be a rich representation of the task itself~\cite{achille2019task2vec}, and the work~\cite{yu2020gradient} addresses the gradient conflicts in multi-task context. The recent work~\cite{Lucas_2021_ICCV} tackles similar gradient conflict caused by the existing different domains in training sets. 
Is it possible to add prior knowledge to guide the gradient for better generalization? According to psychology and neuroscience, the visual perception system of human is shape-biased~\cite{landau1988importance, kucker2019reproducibility}, i.e. relying more on shape information thus owning powerful generlizability. Unfortunately, CNN networks are found to be texture-biased~\cite{ geirhos2018imagenettrained, hermann2020origins}. It is necessary to add shape priori to guide the network for better generalization. 

In this paper, we propose Shape Guided Gradient Voting (SGGV) for domain generalization, which is composed of adding extra inputs to provide shape guidance, and a new gradient updating strategy: gradient voting.
To provide the shape guidance, we have two ways. The first one is to explicitly add edge or sketch of the raw image as an extra shape-distilled input. The second one is through texture augmentation.
The edge or sketch is generated from a traditional Sobel edge operator or a pretrained sketch translating model~\cite{li2019photo}. When the shape-distilled input is fed into the network together with the raw images, the non-conflicting gradients will lead to a shape-biased domain invariant features. Hence the model can be better generalized to unseen domains. 
Besides the explicit shape-distilled input, we also take the texture-augmented input into consideration. 
This is inspired by the observation that the texture augmentation are helpful to learn shape-biased networks~\cite{geirhos2018imagenettrained}.
From the view of image transfer~\cite{huang2017arbitrary,huang2018munit,liu2019few}, the image can be decomposed into content and style. Content is more related with shape while style is more on texture. 
One intuitive understanding is that the various texture augmentation indicates the texture is not the deterministic factor of the semantics, while the consistent shape information is the key to make right recognition. Inspired by this, we also adopted texture-augmented images as a second extra input to provide shape guidance. In a summary, the inputs of our method included the raw images, shape-distilled images (edge/sketch) and texture-augmented images, which are fed into the model simultaneously. This is different from the typical data augmentation setting, which does not require the raw image and augmented one as paired input. Please note that, when generating shape-distilled images, the goal is only to enhance the shape information, and the semantics of these extra inputs may not be well maintained, resulting in some gradient outliers. Next we will discuss how to address this issue with our proposed gradient voting strategy.

Regularly, all the gradients will be used to update the network weights, which is denoted as \emph{Deep-All}. Another option is only non-conflicting gradients are allowed to change the network weights~\cite{Lucas_2021_ICCV,yu2020gradient}. However, neither of these two solutions may be the best choice. The strategy that only allows non-conflicting gradients are sensitive to outliers and tends to prevent the optimization to learn discriminative features even on the training set itself. While Deep-All may overfit the training set, leaving the real generalizable discriminative features hidden under other cluttered features. We provide a balanced solution with a new proposed simple but effective approach, gradient voting. SGGV conducts majority voting on the binary directions for each dimension of the gradients of both the original input and extra inputs for shape guidance, leading to a robust and shape-biased updating direction. 

Our main contribution can be summarised as follows: $(i)$ To the best of our knowledge, we are the first to explicitly introduce shape information as gradient guidance for domain generalization. $(ii)$ We propose a gradient voting method so that the network can learn common knowledge with shape guidance. $(iii)$ We do extensive experiments to verify the effectiveness of our method. 


\section{Related Work}
\label{sec:related}
\paragraph{Domain Generalization}
In the piratical application, there is usually domain gap between the training and testing data, which often leads to performance of the model drop. To address this issue, various methods have been proposed to solve this domain generalization problem~\cite{ijcai2021-628}. The first branch is to learn domain-invariant representation, expecting the unseen target domains can be aligned with the source domain(s)~\cite{muandet2013domain,ganin2016domain,li2018domain,zhao2020domain}.  The second branch is augmentation based methods. Some conduct the augmentation~\cite{volpi2018generalizing,Li_2021_CVPR_progressive,Li_2021_cvpr,zhou2021domain,nuriel2020padain} on images and some on feature space. 
The third one adds or learns knowledge, e.g., including meta-learning for regularization~\cite{balaji2018metareg}, shape prior~\cite{carlucci2019domain,shi2020informative,Narayanan_2021_ICCV}, self-challenging~\cite{huangRSC2020}, Fourier space~\cite{Huang_2021_CVPR,Xu_2021_CVPR}. Other efforts include gradient~\cite{Lucas_2021_ICCV}, model selection~\cite{gulrajani2020search}, ensemble learning~\cite{cha2021swad}, and normalization~\cite{Seo2020DSON,Tang_2021_ICCV,Fan_2021_CVPR}. Among which, the most relevant work to our method is ~\cite{Lucas_2021_ICCV}, which conducts gradient surgery on the gradients and only allow non-conflicting gradients updating. {We have explicit shape prior guidance} and gradient voting updating strategy to neglect outliers. For other papers that leverage shape, we are different from them in our novel way of using the shape as gradient guidance.  
\vspace{-3mm}
\paragraph{Gradient Related}
Gradient descent is almo st the most popular optimization strategy used in machine learning~\cite{ruder2016overview}. The gradients are found to be a good representation of the task~\cite{achille2019task2vec} and used to model the relation of tasks.  In the multi-task learning context, ~\cite{yu2020gradient} addresses the gradient conflicts. In domain generalization literature, the meta-learning based methods~\cite{li2018learning,balaji2018metareg} can also be viewed as a specific operation on the gradient descent. Huang et al.~\cite{huangRSC2020} proposes to discard the representations associated with the higher gradients to leverage more features for prediction. Lucas et al.~\cite{Lucas_2021_ICCV} propose to remove conflicting gradients to find domain invariant features. In our method, the goal is to provide shape prior guidance for the gradient descent. 
\vspace{-3mm}
\paragraph{Data Augmentation}
Data augmentation is a very important and effective technique for machine learning to increase the data diversity. Here we only introduce some augmentation methods related to our method. Regular augmentations for images include scale, random crop, flipping, color jitter~\cite{krizhevsky2012imagenet}.  ~\cite{Cubuk_2019_CVPR} automatically learns data augmentation policies. ~\cite{geirhos2018imagenettrained} finds style augmentation can increase the shape bias of CNN models. Targeting for domain generalization,  ~\cite{volpi2018generalizing, Li_2021_CVPR_progressive} adopt an adaptive data augmentation method to generate adversarial examples. ~\cite{Li_2021_cvpr} studies the augmentation on feature.  ~\cite{zhou2021domain} uses style augmentation to increase the generalization. In this work, we use data augmentation, especially texture augmentation (color jitters) to provide feature mutations for gradient voting. Our method holds the potential to leverage different augmentations to provide feature diversity, in this paper, we only consider the texture related augmentations.
\vspace{-3mm}
\paragraph{Sketch/Edge}
We can extract edge or sketch from images. They provide clean shape information compared to raw images. Traditional edge operators include Laplace, Sobel, Canny and so on. Song et al. ~\cite{ha2018neural} uses LSTM to sequentially generate the stroke for a simple sketch. Le et al. ~\cite{li2019photo} generate sketches with an image translation mechanism. Wang et al.~\cite{wang2021sketchembednet} takes sketch generation as a representation learning task.
In this paper we use the edge/sketch as explicit shape-distilled input for shape guidance in gradient voting.


\begin{figure*}[t]
\centering
\includegraphics[width=1.00\linewidth]{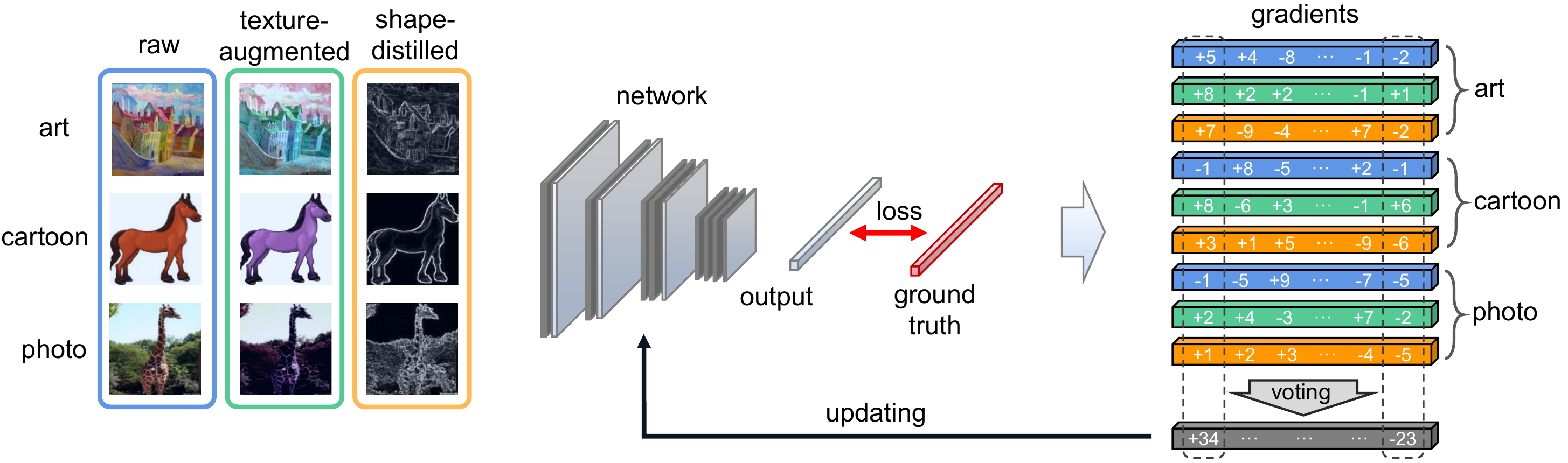}
\caption{An overview of our method. We have texture-augmented (green color) and shape-distilled (orange color) images as extra input to provide shape guidance. We conduct gradient voting to get the final gradient to update the weights of the network.}
\vspace{-5mm}
\label{fig:overview}
\end{figure*}

\section{Shape Guided Gradient Voting}
\label{sec:method}
\subsection{Preliminaries for Domain Generalization}
\label{subsec:Preliminaries for DG}
In the context of domain generalization, the training set consists of $M$ source domains $\mathcal{S}_1, \mathcal{S}_2, ..., \mathcal{S}_M$. The test set is a target domain $\mathcal{T}$, which we can not access during the training stage. For source domain $\mathcal{S}_i$, we have $N_i$ data points $\{(\boldsymbol{x}_j,y_j)\}_{j=1}^{N_i}$, where $\boldsymbol{x}_j$ is the raw image, $y_j$ is the class label. The goal of domain generalization is to learn network $f_{\theta}$ with parameter $\theta$ on source domains $\mathcal{S}_1, \mathcal{S}_2, ..., \mathcal{S}_M$, which can be well generalized to the unseen target domain $\mathcal{T}$. The loss function is the sum of the loss of each domain $\mathcal{L(\theta)} = \sum_{i=1}^M {\mathcal{L}_i(\theta)}$, and the loss function of source domain $\mathcal{S}_i$ is $\mathcal{L}_i(\theta)=\sum_{j=1}^{N_i} {\ell(f_\theta(\boldsymbol{x}_j), y_i)}$, where $\ell(\cdot,\cdot)$ is the classification loss such as cross entropy.
$\boldsymbol{g}_i\in\mathbb{R}^K$ is the gradient vector of $i$-th source domain images $\{\boldsymbol{x}_j\}_{j=1}^{N_i}$ and $\boldsymbol{g}_i=\nabla_\theta \mathcal{L}_i(\theta)$. $K$ is the total number of learnable parameter in the model $f_{\theta}$. $\boldsymbol{g}_i(k)$ is the $k$-th element of $\boldsymbol{g}_i$. In the regular setting, \emph{Deep-All}, the final gradient to update $\theta$ is $\boldsymbol{g} = \sum_{i=1}^M {\boldsymbol{g}_i}$. 

\subsection{Overview}
\label{subsec:Overview}
To achieve better domain generalization ability, we introduce shape priori by adding extra input, when performing gradient updating. There are two main parts in our method, one is the extra input to provide shape guidance, the other is gradient updating strategy. 

The extra inputs include shape-distilled image $\boldsymbol{x}^s_j$ and texture-augmented image $\boldsymbol{x}^t_j$, originated from raw input image $\boldsymbol{x}_j$, as shown in Figure~\ref{fig:overview}. The shape-distilled image $\boldsymbol{x}^s_j$ is an image extracted from raw RGB image as $\boldsymbol{x}^s_j = shape\_dis(\boldsymbol{x}_j)$, which contains less texture information compared to raw image. $shape\_dis(\cdot)$ is a shape distillation operation. The texture-augmented image is the output of texture augmentation operation on raw RGB as $\boldsymbol{x}^t_j = texture\_aug(\boldsymbol{x}_j)$. $texture\_aug(\cdot)$ is a texture augmentation operation, which perturbs the texture of $\boldsymbol{x}_j$. The consistent composition between $\boldsymbol{x}_j$ and $\boldsymbol{x}^t_j$ is likely to be shape-biased. The details of how we get those extra inputs are presented in Section ~\ref{subsec:shape guidance}. 

We let $\boldsymbol{g}_i^s$ and $\boldsymbol{g}_i^t$ denote the gradients of $i$-th domain shape-distilled images $\{\boldsymbol{x}^s_j\}_{j=1}^{N_i}$ and texture augmented images $\{\boldsymbol{x}^t_j\}_{j=1}^{N_i}$, respectively. We would like to make the model better generalized to unseen domain $\mathcal{T}$ with the shape guidance provided by $\boldsymbol{g}_i^s$ and $\boldsymbol{g}_i^t$. The inputs of gradient voting is $\{(\boldsymbol{g}_i,\boldsymbol{g}_i^s,\boldsymbol{g}_i^t)\}_{i=1}^M$, and the final output is a vector $\boldsymbol{g}_{SGGV}\in\mathbb{R}^K$ uesd to update the model weights. 
Since the diversity of features is highly increased with more inputs, meanwhile the outliers are introduced inevitably in the mutation process of shape and texture. As shown in Figure~\ref{fig:overview}, given the $k$-th dimension of gradients $\{(\boldsymbol{g}_i(k),\boldsymbol{g}_i^s(k),\boldsymbol{g}_i^t(k))\}_{i=1}^M$, to find a shape-guided and robust updating direction, we perform a majority voting. The $k$-th element of $\boldsymbol{g}_{SGGV}$ is $\boldsymbol{g}_{SGGV}(k) = voting(\{(\boldsymbol{g}_i(k),\boldsymbol{g}_i^s(k),\boldsymbol{g}_i^t(k))\}_{i=1}^M)$. The $voting(\cdot)$ is a majority voting operation. The details and baselines are presented in Section~\ref{subsec:Gradient voting}.

Our method only modifies the training stage, for testing we only input the raw images from the unseen domain $\mathcal{T}$ for inference, there are no shape-distilled or texture-augmented images as extra input.

\subsection{Shape Guidance}
\label{subsec:shape guidance}
In this section, we present the process how we provide the shape guidance for gradient updating. We adopt an explicit way to enhance the shape information by providing shape enhanced input and an implicit way to emphasis shape through texture augmentation. The first one is to distillate shape information from the raw images and suppress texture composition. Another one is inspired by data augmentation, which disturbs the texture composition of images and leave the shape as the consistent composition. 

The shape-distilled image $\boldsymbol{x}^s_j$ is extracted from a raw RGB image as $\boldsymbol{x}^s_j = shape\_dis(\boldsymbol{x}_j)$. The $shape\_dis(\cdot)$ can be an operator at the pixel level, or a image translation neural network.For the implementation, we select two entities with shape enhanced but texture suppressed: edge and sketch. Edge can remove most of the texture information and highlight the boundaries of objects in the images. Meanwhile, edge can be well obtained through some simple traditional operators, efficiently. Common edge extraction operators include Laplace, Sobel, Canny and so on. Sketch depicts the objects through a composition of strokes. However, it is hard to get a sketch directly from the image with a simple operator. A popular method of obtaining sketches is through neural network~\cite{li2019photo}. Formula of $shape\_dis(\cdot)$ will be described in details in the experiment section. Under the guidance of $\boldsymbol{x}_j^s$, the network can pay more attention on learning shape information, rather than just relying on local texture information. In the experimental section, we will visually show that after adding shape guidance, the network is capable to focus on shape information in the form of global boundary.

In addition to explicitly add distilled shape information, we also implicitly emphasis shape information via texture augmentation. It is well known that the purpose of data augmentation is to reduce the dependence on specific aspects. For example, flipping and rotation aim to reduce the sensitivity of direction, while scaling and cropping attempt to decrease the sensitivity of scale. Similarly, image texture augmentation, such as color transformation, can alleviate the dependence of texture, implicitly making the network focus more on shape information. We generate a texture augmentation version $\boldsymbol{x}_j^t$ from the image $\boldsymbol{x}_j$ as $\boldsymbol{x}^t_j = texture\_aug(\boldsymbol{x}_j)$. $texture\_aug(\cdot)$ represents a series of texture enhancement operations, such as color jitter, blur and so on.

Regular data augmentation uses the augmented images to replace the original images and feed augmented ones to the network. However, our purpose is to provide shape guidance of each sample for better gradient updating. We feed the augmented image $\boldsymbol{x}_j^t$ and raw image $\boldsymbol{x}_j$ to the network in pair, simultaneously, making $\boldsymbol{x}_j^t$ play an implicit role of providing shape guidance.

\subsection{Gradient Voting}
\label{subsec:Gradient voting}
With the provided shape guidance, now we need to leverage it during the gradient updating. 
Following common practice of gradient descent, we adopt a mini-batch updating. For simplicity, we reuse the notions in section~\ref{subsec:Preliminaries for DG} in a mini-batch level for simplicity. 
Specifically, we sample a mini-batch raw images  $\{\boldsymbol{x}_j\}_{j=1}^{B}$, where $B$ is the batch size, their corresponding shape-distilled images $\{\boldsymbol{x}^s_j\}_{j=1}^{B}$ and texture-augmented images $\{\boldsymbol{x}^t_j\}_{j=1}^{B}$, respectively. Then we feed them to the network for forward and back propagation to obtain the gradients $\{(\boldsymbol{g}_i,\boldsymbol{g}_i^s,\boldsymbol{g}_i^t)\}_{i=1}^M$. Finally, we calculate the final gradient $\boldsymbol{g}_{SGGV}$ through voting among $\{(\boldsymbol{g}_i,\boldsymbol{g}_i^s,\boldsymbol{g}_i^t)\}_{i=1}^M$. For simplicity, we rewrite the previous set as $\{\boldsymbol{g}_l\}_{l=1}^{3M}$, with $\boldsymbol{g}_l = \boldsymbol{g}_{l-M}^s$ if $M < l\le 2M$ and $\boldsymbol{g}_l = \boldsymbol{g}_{l-2M}^t$ if $2M < l\le 3M$.

Before we introduce our method, we first give a quick browse of some baselines to obtain the final gradients. Please note that those baseline methods only take the gradients of raw images as input, i.e.,  $\{\boldsymbol{g}_l\}_{l=1}^M$. We can adapt these baselines with the input of shape guidance, i.e., $\{\boldsymbol{g}_l\}_{l=1}^{3M}$. 
The first baseline is a regular solution of gradient decent, Deep-All, $\boldsymbol{g}_{Deep-All} = \sum_{l=1}^{M} {\boldsymbol{g}_l}$. \emph{Agr-Sum} and \emph{Agr-Rand} are proposed in \cite{Lucas_2021_ICCV}, which only allow non-conflict gradient updating. Based on Deep-All, Agr-Sum set $\boldsymbol{g}_{Deep-All}(k)$ to $0$ if the signs of $\{\boldsymbol{g}_l(k)\}_{l=1}^M$ are not the same. To avoid dead weights that are never modified during training, Agr-Rand replace $0$ with a value randomly sampled from a Gaussian distribution.
PCGrad~\cite{yu2020gradient} takes cosine similarity to determine the conflicting case and use projection to remove the conflicting composition. 

Now we present the gradient updating strategy of SGGV. We conduct a majority voting operation on each dimension of the gradient based on the signs of the gradient vectors. Specifically, given a total number of $3M$ gradient vectors, at the $k$-th dimension, suppose there are a number of $m$ positive values and a number of $3M-m$ negative values. We have a hyper-parameter $\tau$ as the threshold for majority voting. If the number of positive values or negative values exceeds the preset threshold $\tau$, the corresponding $k$-th parameter can be modified, otherwise we set the aggregated value to $0$.
More specifically, the $voting(\cdot)$ operation to get the final gradient is calculated as follows: 
\[
\label{equ:sggv}
\begin{split}
\boldsymbol{g}_{SGGV}(k) \!=\!
\left\{
\begin{aligned}
&\sum_{l=1}^{3M}\frac{1+s(\boldsymbol{g}_l(k))}{2}\cdot{\boldsymbol{g}_l(k)}&\!\!\!\!\!\!,\;& i\!f \, m \geq \tau, \\
&\sum_{l=1}^{3M}\frac{1-s(\boldsymbol{g}_l(k))}{2}\cdot{\boldsymbol{g}_l(k)}&\!\!\!\!\!\!,\;& i\!f \, m \leq 3M\!-\!\tau, \\
&\;0&\!\!\!\!\!\!,\;& \; otherwise,
\end{aligned}
\right.
\end{split}
\]

where $s(\cdot)$ is the signum function. 
When the number of positive values $m$ is greater than or equal to the threshold $\tau$, we use the sum of $m$ positive gradient values as the final aggregated value, and discard all negative gradient values at the same time. When the number of positive numbers is smaller than or equal to $3M-\tau$, it means that the number of negative values reaches the threshold $\tau$. Therefore, the sum of all negative gradient values is used as the final aggregated value, and all positive gradient values are discarded. In this way, we let the gradient vectors of the shape-distilled and texture-augmented images join the gradient descent process. By discarding the harmful gradient outliers, we are able to modify the network along a robust and shape guided direction. 
\begin{figure*}[ht]
\centering

\includegraphics[height=.378\textheight]{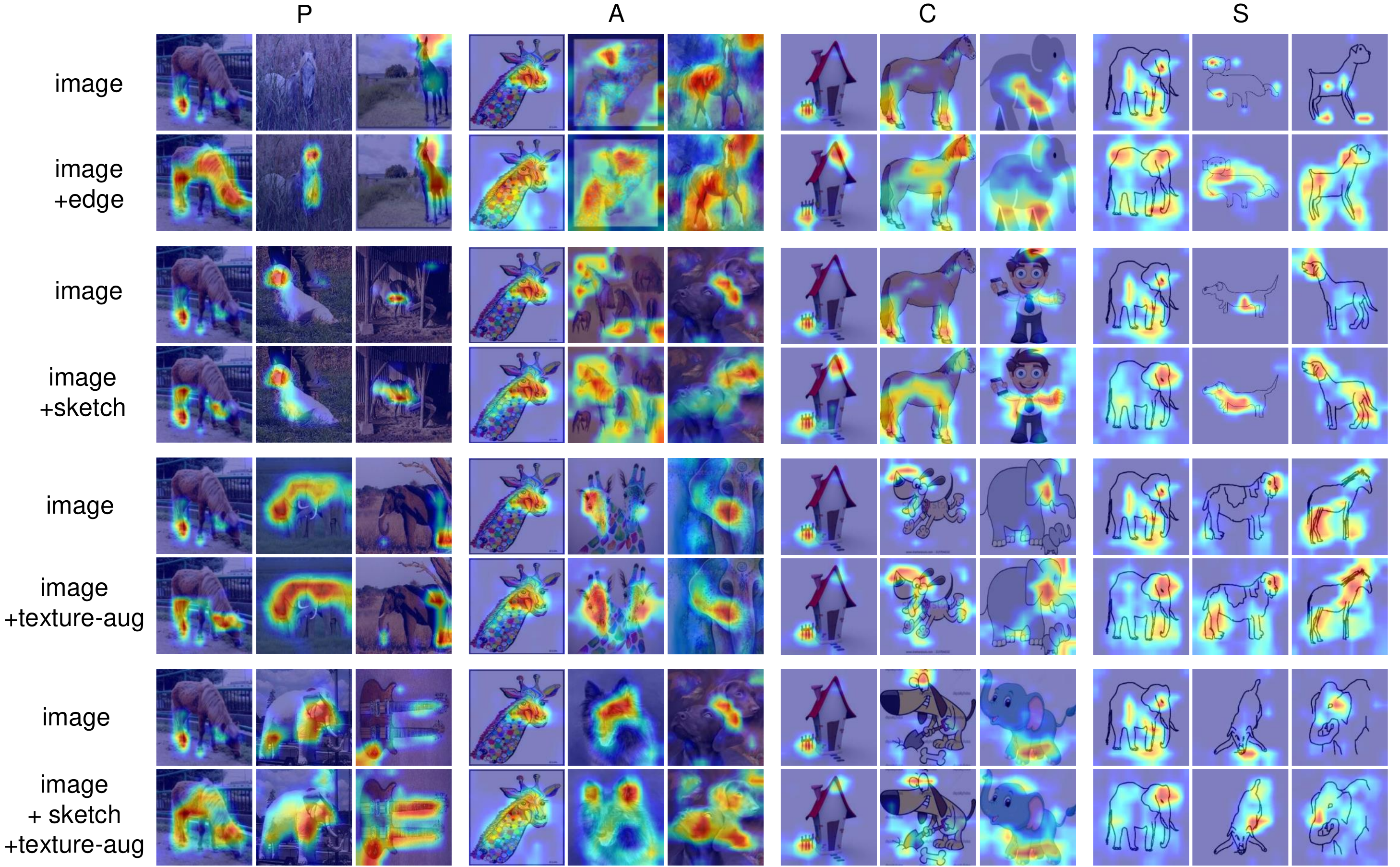}
\vspace{-10pt}
\caption{Visualization of the focused regions using GradCAM. Compared to the image only setting, the focused regions tend to extend to larger region and cover more object boundary after adding shape guidance (image + edge/sketch/texture-aug), especially for S domain. }
\vspace{-6mm}
\label{fig:gradcam}
\end{figure*}

\vspace{-3mm}
\section{Experiment}
\label{sec:exp}
\vspace{-1mm}
\subsection{Experimental Setting}
\label{subsec:setting}
\vspace{-2mm}
\paragraph{\bf{Datasets and implementation details}.} We use three well-known datasets in domain generalization, PACS~\cite{li2017deeper}, VLCS~\cite{fang2013unbiased} and OfficeHome~\cite{venkateswara2017deep}. We conduct classification tasks on these three datasets to verify the effectiveness of our method. To evaluate the generalization on unseen domains, we adopt the leave-one-domain-out strategy following previous work\cite{carlucci2019domain,dou2019domain,li2017deeper,Lucas_2021_ICCV}. 
Following previous work \cite{carlucci2019domain,dou2019domain,li2017deeper,Lucas_2021_ICCV}, we choose Alexnet~\cite{NIPS2012_c399862d} pretrained on Imagenet~\cite{russakovsky2015imagenet} as the backbone for classification tasks.Our SGGV can also be applied to more complex backbones.

\vspace{-2mm}
\paragraph{\bf{Baselines of Gradient Updating Strategy}.}
Here by the gradient updating strategy, we mean the operation on the gradients of different domains or tasks, rather than the techniques designed for general optimization such as Adam~\cite{kingma2015adam}. Here we choose the baselines as we discussed in section~\ref{subsec:Gradient voting}: Deep-All, Agr-Sum, Agr-Rand and PCGrad. For fair comparison, we rerun them and adapt them when we have the shape guidance input by treating the shape-distilled and texture-augmented as the new domains.

\vspace{-3mm}
\begin{table*}[ht!]
\centering
\setlength{\tabcolsep}{3mm}
\caption{Ablation study on PACS. The numbers are the accuracy in percentage. The best results inside each block are underlined. The best ones in the table are in bold. The ~\emph{input} column illustrates the input types we use during training.  For example, image+edge indicates that images and edges are input into the network together.}
\begin{tabular}{c c c c c c c c} 
 \hline
 method & input & A & C & P & S & Avg.\\  
 \hline
 Agr-Sum & image  & 63.49 & 64.88 & 85.33 & 58.87 & 68.14 \\
 Deep-All & image  & 60.22 & 62.84 & 84.73 & 59.06 & 66.71 \\
 \hline
 \rowcolor{pink} Agr-Sum & image+edge  & \underline{64.85} & 63.82 & 84.34 & 58.61 & 67.91 \\ 
 \rowcolor{pink} Deep-All & image+edge  & 61.46 & 63.33 & 83.95 & \underline{64.35} & 68.27 \\
 \rowcolor{pink} SGGV & image+edge  & 63.71 & \underline{65.37} & \underline{85.33} & 63.84 & \underline{69.56} \\ 
 \hline
  \rowcolor{green} Agr-Sum & image+sketch  & 62.88 & 66.48 & 86.08 & 66.36 & 70.45 \\ 
  \rowcolor{green} Deep-All & image+sketch  & 60.39 & 63.88 & 84.40 & \underline{68.49} & 69.29 \\
  \rowcolor{green} SGGV & image+sketch  & \underline{64.00} & \underline{66.16} & \underline{86.62} & 67.81 & \underline{71.15} \\
 \hline
  \rowcolor{yellow!70} Agr-Sum & image+texture-aug  & 66.05 & 64.48 & 85.72 & 60.82 & 69.27 \\ 
  \rowcolor{yellow!70} Deep-All & image+texture-aug  & 64.83 & 62.50 & 85.51 & \underline{66.03} & 69.72 \\
  \rowcolor{yellow!70} SGGV & image+texture-aug  & \underline{68.12} & \underline{65.48} & \underline{87.13} & 63.35 & \underline{71.02} \\
 \hline
 \rowcolor{gray!50} Agr-Sum & image+sketch+texture-aug  & 65.68 & 66.16 & 86.02 & 67.58 & 71.36 \\
 \rowcolor{gray!50} Deep-All & image+sketch+texture-aug  & 64.29 & 64.43 & 85.90 & \underline{\textbf{72.16}} & 71.70 \\
 \rowcolor{gray!50} SGGV & image+sketch+texture-aug  & \underline{\textbf{69.00}} & \underline{\textbf{66.50}} & \underline{\textbf{87.43}} & 70.42 & \underline{\textbf{73.34}} \\
 \hline
\end{tabular}
\label{table:ablation_shape}
\end{table*}

\vspace{-5mm}
\subsection{Ablation study on Shape Guidance}
\label{subsec:ablation_shape}
\vspace{-2mm}
In this section, we conduct ablation study on PACS to understand how our SGGV works. For the shape guidance part, we study the impact of shape-distilled images (edge and sketch), texture-augmented images individually. Then we put them together to fully provide shape guidance. 

To get the shape-distilled images, we adopt Sobel operator as the $shape\_dis(\cdot)$ for edge. For sketch, we adopt a pretrained translation model from \cite{li2019photo} as the $shape\_dis(\cdot)$. For texture-augmented image, we use color jitter as $texture\_aug(\cdot)$. The parameters of color jitter are set as: brightness to 0.8, contrast to 0.8, saturation to 0.8, and hue to 0.5. The results are shown in Table \ref{table:ablation_shape}. The results using edge as extra input are labelled with pink background. The results of sketch are in green, and the results of texture-augmented images are in yellow. The grey block shows the results of a full version of shape guidance with both explicit shape-distilled images (sketch) and implicit texture-augmented images to provide shape guidance. 
In this section, since our SGGV is designed for the situation with shape guidance, we only focus on the results of the baselines of Agr-Sum and Deep-All. 

\vspace{-2mm}
\paragraph{\bf{Quantitative Results}.}
For the result of edge as extra input, if we do not discard the conflicting gradients and modify all the network parameters, Deep-All with image+edge as inputs gets average accuracy of 68.27\%, which is higher than Deep-All with image as input (66.71\%). This indicates the extra shape input results in better generalization ability. However, with image+edge as input, Arg-Sum (67.91\%) is worse than that without edge(68.14\%). We suppose the reason is that Arg-Sum is sensitive to the outliers introduce by edge, and thus making the gradient conflicting get severely.  
For sketch as extra input, both Agr-Sum and Deep-All gain a lot than the result with only image input(2.31\% and 2.58\%, respectively) and images with edges(2.54\% and 1.02\%, respectively). 
The reason is supposed to be that, on PACS dataset, the sketch contains more precise shape information than edge and provides better shape guidance.
For texture-augmented images as input, both Agr-Sum and Deep-All also outperform the ones with only image input.
By merging the explicit and implicit shape guidance together, the performance of Agr-Sum and Deep-All is further improved 3.22\% and 4.99\% respectively, indicating  shape-distilled and texture-augmented images can work together complementarily.
\vspace{-3mm}
\paragraph{\bf{Qualitative Results}.}
We also provide the qualitative result by using Grad-CAM \cite{selvaraju2017grad} to visualize the changes of the focused regions before and after adding shape guidance. The result is shown in Figure~\ref{fig:gradcam}.

After explicitly adding shape guidance, the focused regions of the network expand to the boundary of the object, especially on sketch domain. The visualization verifies our idea that the explicit addition of shape information really plays a guiding role for the network, so that the network can pay more attention to grasp shape information, which is beneficial for generalization.  Another interesting observation is that, for texture-augmented image, the expanding region is mainly around the original region. While, for the shape-distilled image, the new focused region tends to cover the boundary. Two different expanding phenomena actually indicate that these two types of shape guidance are complementary and can work together.

\vspace{-2mm}
\begin{table}[ht!]
\centering
\setlength{\tabcolsep}{1.3mm}
\begin{minipage}[t]{0.47\textwidth}
\caption{Gradient voting without shape guidance to evaluate the importance of shape guidance. We do not provide any shape guidance, only use raw image as input. }
\setlength{\tabcolsep}{1.0mm}
\begin{tabular}{c c c c c c} 
 \hline
 method & A & C & P & S & Avg\\  
 \hline
 Agr-Sum & 63.49 & 64.88 & 85.33 & 58.87 & 68.14 \\ 
 SGGV-2 & 62.46 & 63.58 & 85.66 & 61.06 & 68.19 \\
 \hline
\end{tabular}
\label{table:noshapeguide}
\end{minipage}
\hspace{1mm}
\begin{minipage}[t]{0.5\textwidth}
\setlength{\tabcolsep}{1.2mm}
\caption{We count the proportion of retained elements of gradients with and without shape guidance. The total number of elements is about 57M. All numbers in the table are percentages.}
\begin{tabular}{c c c c c c} 
 \hline
 method & A & C & P & S & Avg\\  
 \hline
 w/o shape & 66.06 & 65.64 & 66.78 & 69.38 & 66.97 \\
 w shape & 79.03 & 77.69 & 79.41 & 84.35 & 80.12 \\
 \hline
\end{tabular}
\label{table:Similarity}
\end{minipage}
\end{table}

\vspace{-4mm}
\subsection{Ablation study on Gradient Voting}
\label{subsec:ablatioon_shape}
\vspace{-2mm}
\paragraph{\bf{Types of Shape Guidance}.}
For SGGV, we first study the impact of different types of shape guidance by comparing to other baselines. The voting threshold we use are $\tau$=$2L/3$, where $L$ is the total number of the input gradient vectors. For image + edge/sketch/texture-aug, $\tau$=$4$ and for image + sketch + texture-aug, $\tau$=$6$. The result is shown in Table~\ref{table:ablation_shape}. Our SGGV outperforms both Arg-Sum and Deep-All on all types of shape guidance. For edge, SGGV (69.56\%) surpasses both Deep-All (68.27\%) and Arg-Sum (67.91\%) with a large gap, which validate that our proposed method can better leverage the shape information provided by the extra edge input. Let's zoom in to have a look at each specific domains. Surprisingly, the gain is mainly from A,C,P domains, even the edge is much closer to the S domain in appearance. This observation further illustrates that SGGV leads to more attention on shape, rather than only adding more data. 
For sketch, SGGV (71.15\%) also surpasses both Deep-All (69.29\%) and Arg-Sum (70.45\%). This result is quite consistent with the result of edge, which supports the analysis above. 
For texture-aug, SGGV (71.02\%) still outperforms Deep-All (69.72\%) and Arg-Sum (69.27\%). The gap between Deep-All and SGGV is quite smaller than the green and pink ones. This is because the color jitter is a very basic texture augmentation, there is small chance to generate outliers, and modifying all the parameters can also lead to a good solution.

The above experiment shows that the gradient voting can work well with different types of shape guidance. We conduct the following experiment to evaluate the importance of shape guidance. 
We do not provide any shape guidance, only use raw image as input. We modify the threshold of gradient voting strategy as $\tau=2$. The result is shown in Table~\ref{table:noshapeguide}.  It can be seen that without the shape guidance, the gradient voting gets 68.19\%, which is quite similar with Agr-Sum (68.14\%). Therefore, shape guidance is necessary for seeking generalizable features.

\vspace{-2mm}
\paragraph{\bf{SGGV increases gradient similarity}.}
We do an ablation study that demonstrates how taking into consideration the extended inputs with the proposed gradient voting strategy helps to select a gradient that leads to a model which is better capable to generalize. The results are shown in table \ref{table:Similarity}. 
We find that the proportion of retained elements of gradients increases a lot with shape guidance. Hence the selected gradient tends to find the invariant characteristics among different domains. Finally, the model is able to learn domain-invariant features and prefers to grasp global shape information. So the model is better capable to generalize.

\vspace{-2mm}
\paragraph{\bf{Using paired inputs}.}
In table \ref{table:in pair}, we show that how images and sketches pair or not impacts the performance. We find that gradient methods(Agr-sum and SGGV) gain a lot through paired inputs while Deep-All not. For some dimensions of gradients, using unpaired inputs may introduce some extra gradient collisions so that causing useful gradients to be zeroed. For example, round heads of animals and sharp corners in houses probably are significant clues for network to classify according to Grad-CAM. The elements in gradients denoting these two clues may be discarded when they occur at same dimension in gradient vector if inputs are unpaired. But the clues will be contained in pairing circumstance.

\begin{table}[t]
\centering
\setlength{\tabcolsep}{2.1mm}
\begin{minipage}[t]{0.4\textwidth}
\caption{The impact of voting threshold $\tau$ on the number of model parameters that can be modified. S.V. means SGGV.}
\begin{tabular}{c c c c } 
\hline
 method & S.V.-5 & S.V.-4 & S.V.-3 \\  
 \hline
 A & \textbf{64.37} & 63.71 & 63.10 \\
 C & 65.35 & \textbf{65.37} & 64.84 \\ 
 P & 85.29 & 85.33 & \textbf{85.60}\\
 S & 62.53 & 63.84 & \textbf{64.41}\\
 Avg & 69.39 & \textbf{69.56} & 69.49\\
 para/M & 12$\sim$16 & 37$\sim$41 & 57\\
 \hline
\end{tabular}

\label{table:param}
\end{minipage}
\hspace{2mm}
\setlength{\tabcolsep}{1.5mm}
\begin{minipage}[t]{0.55\textwidth}
\caption{Different methods with paired inputs or not on PACS. SGGV uses paired inputs and $\tau$=5. A.S. means Agr-Sum and D.A. means Deep-All.}
\vspace{1mm}
\begin{tabular}{c c c c c c c} 
 \hline
 method & pair & A & C & P & S & Avg\\  
 \hline
 A.S. & no & 62.12 & 65.61 & 84.28 & 66.09 & 69.53 \\ 
 D.A. & no & 61.02 & 64.33 & 84.85 & \textbf{68.64} & 69.71 \\
 \hline
 A.S. & yes & 62.88 & 66.48 & 86.08 & 66.36 & 70.45 \\ 
 D.A. & yes & 60.39 & 63.88 & 84.40 & 68.49 & 69.29 \\
 SGGV & yes & \textbf{65.32} & \textbf{66.57} & \textbf{86.86} & 67.38 & \textbf{71.53} \\ 
 \hline
\end{tabular}
\label{table:in pair}
\end{minipage}
\vspace{-3mm}
\end{table}

\vspace{-2mm}
\paragraph{\bf{Voting Threshold}.}
Here we do the ablation on the voting threshold $\tau$. We conduct experiments with different $\tau$ on different types of shape guidance. 
We denote SGGV-4 as SGGV with voting threshold $\tau=4$, and the same with SGGV-5, SGGV-6, SGGV-7. The result is shown in Table~\ref{table:ablation_threshold}. 
The observation is that SGGV with different thresholds have quite similar performance on all types of shape guidance, which indicates that this voting strategy is not sensitive to $\tau$. To better understand how the threshold impact the optimization process, we further conduct an experiment to calculate how many parameters are updated with different settings of $\tau$. The experiment takes edge as shape guidance. The result is shown in Table~\ref{table:param}. Since we have 4 different settings of source domains for training, we report the range of modifiable parameters in those 4 cases. As the threshold $\tau$ decreases, the number of parameters that can be modified increases gradually. However, the performance of the $\tau=3$ row drops compared to the one of $\tau=4$. Although more parameters can be updated, real generalizable discriminating features may be overwhelmed with other clustered ones. When $\tau$ is large, fewer parameters can be modified, preventing the network to learn discriminatory features during training.

\begin{table}[t]
\centering
\renewcommand\tabcolsep{3mm}
\caption{Ablation study of the voting threshold $\tau$. i: image, e:edge, s:sketch, t: texture\_aug. SGGV-5 and SGGV-4 (with image+sketch/edge/texture\_aug as input) indicates our methods with threshold $\tau=5$ or $4$, respectively. SGGV-7 and SGGV-6 (with image+sketch+texture-aug as input) indicates our methods with threshold $\tau=7$ or $6$, respectively.}
\begin{tabular}{c c c c c c c c} 
 \hline
 method & input & A & C & P & S & Avg.\\  
\hline
SGGV-5 & i+e  & {64.37} & 65.35 & 85.29 & 62.53 & 69.39 \\
SGGV-4 & i+e  & 63.71 & {65.37} & {85.33} & 63.84 & {69.56} \\ 
\hline

SGGV-5 & i+s  & {65.32} & {66.57} & {86.86} & 67.38 & {71.53} \\
SGGV-4 & i+s  & 64.00 & 66.16 & 86.62 & 67.81 & 71.15 \\
\hline
SGGV-5 & i+t  & 67.75 & {65.65} & {86.96} & 62.82 & 70.90 \\
SGGV-4 & i+t  & {68.12} & 65.48 & 87.13 & {63.35} & {71.02} \\
 \hline
SGGV-7 & i+s+t  & {{69.07}} & {{66.52}} & 87.19 & 69.61 & 73.10 \\
SGGV-6 & i+s+t  & 69.00 & 66.50 & {{87.43}} & 70.42 & {{73.34}} \\
 \hline
\end{tabular}
\vspace{-4mm}
\label{table:ablation_threshold}
\end{table}

\subsection{Comparing with the SOTA methods}
\label{subsec:SOTA}
\vspace{-2mm}
In this section, we compare SGGV to other SOTA methods. SGGV takes original images, texture-augmented images and shape-distilled images together as the shape guidance. We set the threshold $\tau$=$6$ for all the datasets. 
For the shape-distilled images, we select edge or sketch depending on the quality on specific dataset empirically. 
We select sketch as shape-distilled image for PACS and edge for VLCS and OfficeHome.

We compare our proposed method with baseline model Deep-All and other methods including Agr-Sum and Agr-Rand\cite{Lucas_2021_ICCV}, PCGrad \cite{yu2020gradient}, IRM (Invariant Risk Minimization) \cite{arjovsky2019invariant}, MLDG (Meta-Learning Domain Generalization) \cite{li2018learning}, Mixup (Inter-domain Mixup) \cite{yan2020improve}, DRO (Group Distributionally Robust Optimization) \cite{sagawa2019distributionally}. 

The results are shown in table \ref{table:comparison}.
Compared to other methods, Our SGGV has a significant performance improvement on PACS,VLCS and OfficeHome on all target domains, which verifies the effectiveness of our method. Under the 4 settings on PACS, our method has performance growth ranging from 2.70\% to 11.36\% compared to Deep-All. Average accuracy of SGGV is also the best among SOTA methods. There are similar observations on VLCS and OfficeHome.

\begin{table*}[t]
\centering
\setlength{\tabcolsep}{1.2mm}
\caption{Comparison with other methods on PACS, VLCS and OfficeHome.The best one in the table is in bold.}
\vspace{2mm}
\begin{tabular}{c c c c c c c c c c c} 
 \hline
 &&&&&PACS&&&& \\
 \hline
 \scriptsize Source & \scriptsize Target & \scriptsize SGGV & \scriptsize Deep-All & \scriptsize Agr-Sum & \scriptsize Agr-Rand & \scriptsize PCGrad & \scriptsize IRM & \scriptsize MLDG & \scriptsize Mixup & \scriptsize DRO\\  
 \hline
 C,P,S & A & \textbf{69.00} & 60.22 & 63.49 & 61.93 & 61.22 & 54.59 & 55.88 & 55.88 & 54.96 \\ 
 A,P,S & C & \textbf{66.50} & 62.84 & 64.88 & 64.39 & 62.10 & 57.72 & 57.99 & 58.08 & 58.36 \\
 A,C,S & P & \textbf{87.43} & 84.73 & 85.33 & 84.88 & 84.76 & 86.30 & 86.63 & 84.55 & 86.63 \\
 A,C,P & S & \textbf{70.42} & 59.06 & 58.87 & 56.86 & 59.35 & 53.86 & 55.18 & 50.81 & 53.21 \\
     & Avg. & \textbf{73.34} & 66.71 & 68.14 & 67.02 & 66.86 & 63.12 & 63.92 & 62.33 & 63.29 \\ 
 \hline
  &&&&&VLCS&&&& \\
 \hline
 \scriptsize Source & \scriptsize Target & \scriptsize SGGV & \scriptsize Deep-All & \scriptsize Agr-Sum & \scriptsize Agr-Rand & \scriptsize PCGrad & \scriptsize IRM & \scriptsize MLDG & \scriptsize Mixup & \scriptsize DRO\\  
 \hline
 L,S,V & C & \textbf{94.57} & 93.22 & 92.37 & 91.66 & 93.08 & 93.29 & 93.18 & 92.54 & 92.44 \\ 
 C,S,V & L & 57.49 & 57.41 & \textbf{57.56} & 56.45 & 56.47 & 59.22 & 58.55 & 59.02 & 58.40 \\
 C,L,V & S & \textbf{67.06} & 64.44 & 60.72 & 61.45 & 65.08 & 64.16 & 64.11 & 64.98 & 64.11 \\
 A,C,P & V & 67.40 & 65.87 & 62.78 & 63.08 & 66.21 & 67.57 & 67.10 & \textbf{67.68} & 67.08 \\
     & Avg. & \textbf{71.63} & 70.24 & 68.36 & 68.16 & 70.21 & 71.06 & 70.74 & 71.06 & 70.51 \\ 
 \hline
  &&&&&OfficeHome&&&& \\
 \hline
 \scriptsize Source & \scriptsize Target & \scriptsize SGGV & \scriptsize Deep-All & \scriptsize Agr-Sum & \scriptsize Agr-Rand & \scriptsize PCGrad & \scriptsize IRM & \scriptsize MLDG & \scriptsize Mixup & \scriptsize DRO\\   
 \hline
 C,P,R & A & \textbf{36.58} & 32.86 & 35.42 & 35.71 & 32.74 & 33.07 & 33.73 & 35.69 & 33.25 \\ 
 A,P,R & C & \textbf{37.43} & 31.92 & 32.97 & 32.68 & 31.71 & 34.34 & 35.10 & 35.74 & 35.27 \\
 A,C,R & P & \textbf{56.01} & 52.65 & 54.51 & 53.73 & 54.09 & 52.16 & 54.85 & 55.20 & 54.28 \\
 A,C,P & R & \textbf{60.00} & 57.78 & 59.53 & 57.90 & 57.65 & 54.81 & 56.27 & 57.33 & 55.84 \\
     & Avg. & \textbf{47.51} & 43.80 & 45.61 & 45.01 & 44.05 & 43.59 & 44.99 & 45.99 & 44.66 \\
 \hline
\end{tabular}
\label{table:comparison}
\vspace{-5mm}
\end{table*}

\vspace{-3mm}
\section{Conclusion}
\label{sec:Conclusion}
\vspace{-3mm}
In this work, we propose to introduce shape information to guide network learning in gradient modification. We also propose a gradient voting operation to make better use of these shape information. Our method provides a constructive way to better leverage shape priori from the view of gradient decent. The effectiveness of our method is validated with extensive experiments. 

For the limitation, since we adopt an  explicit way to insert shape information, the performance of SGGV is affected by the quality of shape-distilled images. In the future work, we will study how to integrate shape information in a better way, so as to further improve the classification accuracy in domain generalization task. For example, one possible direction is to take the shape extraction and feature learning into an end to end training framework. In addition, our SGGV method can also be extended to work with other types of data augmentation aiming for different kinds of prior knowledge.

{\small
\bibliographystyle{plain}  
\bibliography{references}
}


\end{document}


\maketitle

\section{Implementation Details}

\paragraph{\bf{Datasets}.} We use three well-known datasets in domain generalization, PACS~\cite{li2017deeper}, VLCS~\cite{fang2013unbiased} and OfficeHome~\cite{venkateswara2017deep}. We conduct classification tasks on these three datasets to verify the effectiveness of our method. There 
are 9991 images in PACS, which is composed of four domains: Art (A), cartoon (C), photo (P) and sketch (S), each of which has seven categories. VLCS contains 10729 images. Four domains of VLCS are Caltech101 (C), Labelme (L), Sun09 (S) and Voc2007 (V). Each domain has five categories. There are 15588 images in OfficeHome, which include 65 categories, are composed of four domains: Art (A), Clipart (C), Product (P) and Real World (R). Due to the different image sizes of each dataset, we crop them to ensure that all images of three datasets have the same size of $227\times227$. 
To evaluate the generalization on unseen domains, we adopt the leave-one-domain-out strategy following previous work\cite{carlucci2019domain,dou2019domain,li2017deeper,Lucas_2021_ICCV}. 
We follow the same setting as~\cite{Lucas_2021_ICCV} to randomly split each domain into training, validation and testing subsets as the ratio of $70\%$, $10\%$, $20\%$ respectively.


\paragraph{\bf{Baselines of Gradient Updating Strategy}.}
Here by the gradient updating strategy, we mean the operation on the gradients of different domains or tasks, rather than the techniques designed for general optimization such as Adam~\cite{kingma2015adam}. Here we choose the baselines as we discussed in section 3.4 in paper: Deep-All, Agr-Sum, Agr-Rand and PCGrad. For fair comparison, we rerun them and adapt them when we have the shape guidance input by treating the shape-distilled and texture-augmented as the new domains. For example, there are 3 domains for training on PACS, with shape-distilled and texture-augmented images as extra inputs, we have 9 domains for those baselines.

\paragraph{\bf{Hyper-parameters}.} For the hyper-parameters of classification, we follow the hyper-parameter setting of \cite{Lucas_2021_ICCV} for fair comparison. We choose cross-entropy loss as optimized objective. Learning rate is $1e$-$5$ and batch size is 128 for all experiments. Then we use Adam optimizer and adopt weight decay $5e$-$5$ as regularization strategy. Finally we train all the methods for 1000 iterations and validate every 20 iterations. 
To select model, we follow the same setting with \cite{Lucas_2021_ICCV,gulrajani2020search}, i.e. selecting the model based on the source domain validation set. 
To avoid the randomness caused by random seeding, we report the mean of the results of ten runs with different random seeds.
For the voting threshold $\tau$, we will conduct ablation study in the following experiment.

\section{Difference compared to typical data augmentation}
The difference is three-fold: $(i)$ in terms of the input, typical data augmentation conducts transformations on raw image and only the augmented one is fed into the model. For SGGV, however, the raw image, paired texture augmented one and shape-distilled one are all fed into the model meanwhile. $(ii)$ In terms of the gradients updating, SGGV takes the gradient voting strategy to get the major components. For typical data augmentation, however, there is no specific design for domain generalization task. $(iii)$ In terms of the motivation, typical data augmentation aims to provide more diverse input. For SGGV, however, the goal of extra paired texture augmented and shape-distilled images is to provide explicit shape guidance of the raw image. To our best knowledge, SGGV is the first one to propose edge or sketch as shape-biased transformation for domain generalization task. We also conduct an experiment to verify that SGGV achieves better generalization ability than typical data augmentation as Table~\ref{table:Typical augmentation} shows.

\begin{table}[ht!]
\centering
\setlength{\tabcolsep}{5mm}

\caption{Typical augmentation only chooses augmented images as inputs. SGGV feeds three kinds of image versions to network. "Aug." means typical augmentation.}
\begin{tabular}{c c c c c c} 
 \hline
 method & A & C & P & S & Avg.\\  
 \hline
 Aug. & 63.05 & 63.47 & 85.21 & \textbf{71.95} & 70.92 \\ 
 SGGV & \textbf{69.00} & \textbf{66.50} & \textbf{87.43} & 70.42 & \textbf{73.34} \\
 \hline
\end{tabular}
\label{table:Typical augmentation}

\end{table}

{\small
\bibliographystyle{plain}  
\bibliography{references}
}